\documentclass{article}
\usepackage{ijcai20}

\usepackage{times}

\usepackage{soul}
\usepackage{url}
\usepackage[hidelinks]{hyperref}
\usepackage[utf8]{inputenc}
\usepackage[small]{caption}
\usepackage{graphicx}
\usepackage{amsmath}
\usepackage{booktabs, siunitx, multirow}
\usepackage{dsfont}
\usepackage{color}
\usepackage{balance}
\usepackage{amssymb}

\usepackage{amsmath,amsfonts,bm}

\def\eqref#1{equation~\ref{#1}}

\def\1{\bm{1}}

\DeclareMathAlphabet{\mathsfit}{\encodingdefault}{\sfdefault}{m}{sl}
\SetMathAlphabet{\mathsfit}{bold}{\encodingdefault}{\sfdefault}{bx}{n}

\newcommand{\R}{\mathbb{R}}

\def\sub#1{_{\rm #1}}

\def\eg{{\it e.g.}}

\def\ie{{\it i.e.}}

\title{MULTIPOLAR: Multi-Source Policy Aggregation for Transfer Reinforcement Learning between Diverse Environmental Dynamics}

\author{
Mohammadamin Barekatain$^{1,2}$\footnote{Work done as an intern at OMRON SINIC X}\and
Ryo Yonetani$^1$\And
Masashi Hamaya$^1$\\
\affiliations
$^1$OMRON SINIC X\\
$^2$Technical University of Munich\\
\emails
m.barekatain@tum.de,
\{ryo.yonetani, masashi.hamaya\}@sinicx.com,
}

\begin{document}

\maketitle

\begin{abstract}
Transfer reinforcement learning (RL) aims at improving the learning efficiency of an agent by exploiting knowledge from other source agents trained on relevant tasks. However, it remains challenging to transfer knowledge between different environmental dynamics without having access to the source environments. In this work, we explore a new challenge in transfer RL, where only a set of source policies collected under diverse unknown dynamics is available for learning a target task efficiently. To address this problem, the proposed approach, MULTI-source POLicy AggRegation (MULTIPOLAR), comprises two key techniques. We learn to aggregate the actions provided by the source policies adaptively to maximize the target task performance. Meanwhile, we learn an auxiliary network that predicts residuals around the aggregated actions, which ensures the target policy's expressiveness even when some of the source policies perform poorly. We demonstrated the effectiveness of MULTIPOLAR through an extensive experimental evaluation across six simulated environments ranging from classic control problems to challenging robotics simulations, under both continuous and discrete action spaces. The demo videos and code are available on the project webpage: \url{https://omron-sinicx.github.io/multipolar/}.
\end{abstract}

\section{Introduction}
\label{sec:intro}

We envision a future scenario where a variety of robotic systems, which are each trained or manually engineered to solve a similar task, provide their policies for a new robot to learn a relevant task quickly. For example, imagine various pick-and-place robots working in factories all over the world. Depending on the manufacturer, these robots will differ in their kinematics (\eg, link length, joint orientation) and dynamics (\eg, link mass, joint damping, friction, inertia). They could provide their policies to a new robot~\cite{devin2017learning}, even though their dynamics factors, on which the policies are implicitly conditioned, are not typically available~\cite{Chen2018}. Moreover, we cannot rely on a history of their individual experiences, as they may be unavailable due to a lack of communication between factories or prohibitively large dataset sizes. In such scenarios, we argue that a key technique to develop is the ability to transfer knowledge from a collection of robots to a new robot quickly \emph{only by exploiting their policies while being agnostic to their different kinematics and dynamics}, rather than collecting a vast amount of samples to train the new robot from scratch.

\begin{figure}[t]
  \begin{center}
    \includegraphics[width=.9\linewidth]{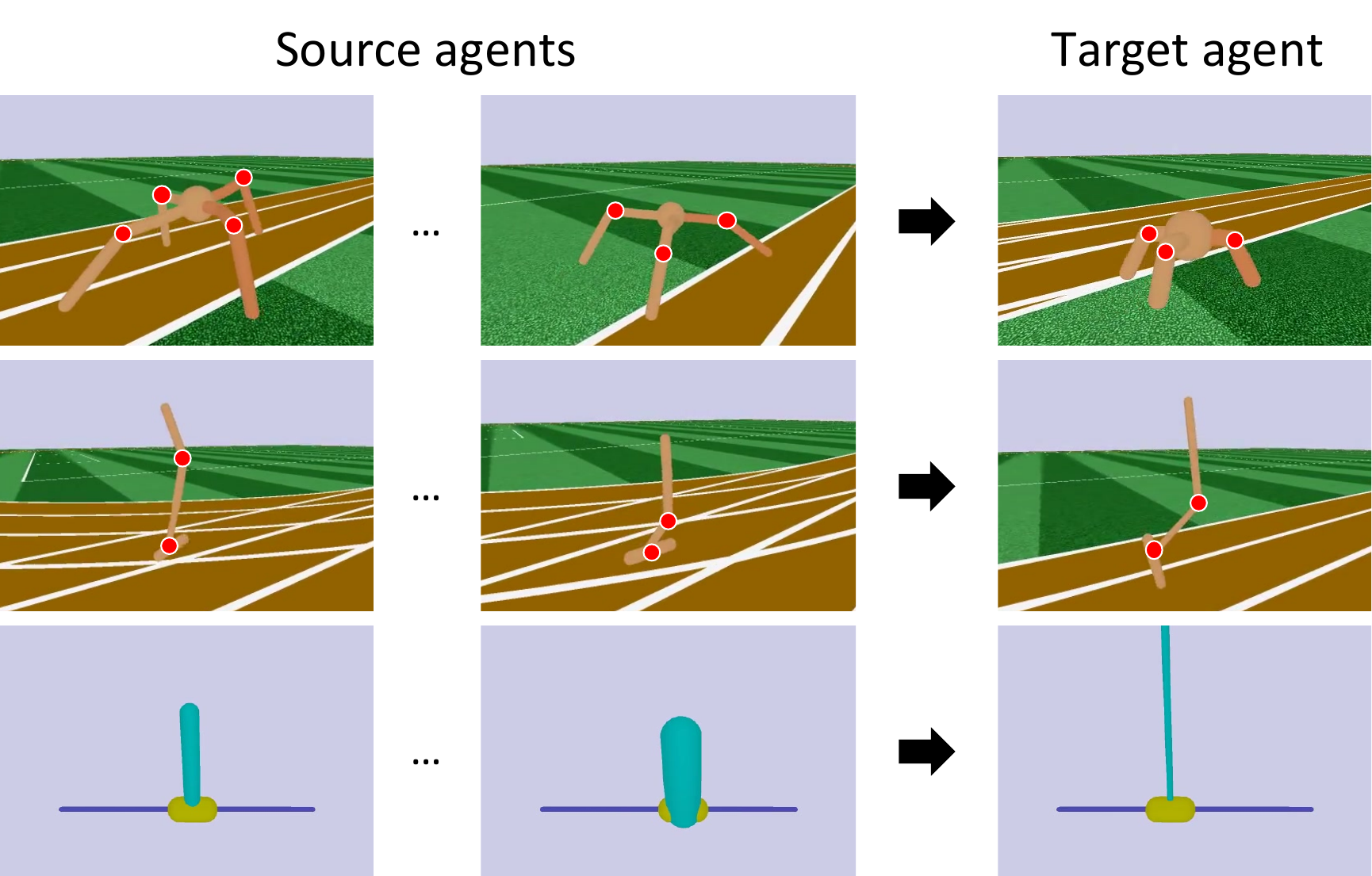}
    \caption{Transfer RL between diverse environmental dynamics. We aim to train a policy of a target agent efficiently \emph{by utilizing the policies of other source agents} under different unknown environmental dynamics. Some joints are annotated by red circles to highlight kinematic diversity.}
    \label{fig:example}
  \end{center}
\end{figure}
The scenario illustrated above poses a new challenge in the transfer learning for reinforcement learning (RL) domains. Formally, consider multiple instances of a single environment with diverse state transition dynamics, \eg, independent robots presented in Figure~\ref{fig:example}, which reach different states by executing the same actions due to the differences in their kinematics and dynamics designs. Some source agents interacting with one of the environment instances provide their deterministic policy to a new target agent in another environment instance. Then, our problem is: \emph{can we efficiently learn the policy of a target agent given only the collection of source policies?} Note that information about source environmental dynamics, such as the exact state transition distributions and the history of environmental states, will not be visible to the target agent as mentioned above. Also, the source policies are neither trained nor hand-engineered for the target environment instance, and therefore not guaranteed to work optimally and may even fail~\cite{Chen2018}. Importantly, these conditions prevent us from adopting existing works on transfer RL between different environmental dynamics, as they require access to source environment instances or their dynamics for training a target policy~(\eg, \cite{Lazaric2008,Chen2018,yu2018policy,pmlr-v80-tirinzoni18a,parisotto:actormimic}). Similarly, meta-learning approaches such as \cite{metarl,clavera2018learning}, cannot be used here because they typically train an agent on a diverse set of tasks (\ie, environment instances). Also, existing techniques that utilize a collection of source policies, \eg, policy reuse frameworks~\cite{fernandez2006probabilistic,rosman2016bayesian,zheng2018deep} and option frameworks~\cite{sutton1999between,Bacon2016,mankowitz2018learning}, are not a promising solution because, to our knowledge, they assume that source policies have the same environmental dynamics but different goals. The most relevant work is ``attend, adapt, and transfer'' (A2T) method~\cite{rajendran:attend} that learns to attend multiple source policies to enable selective transfer. However, this method assumes access to the action probability distribution of source policies (\ie, stochastic source policies), making it hard to address our problem where only deterministic actions of source policies are available (\ie, deterministic source policies). Besides, their presented method is tested only in environments with a discrete action space.

\begin{figure*}[t]
\centering
\includegraphics[width=.9\linewidth]{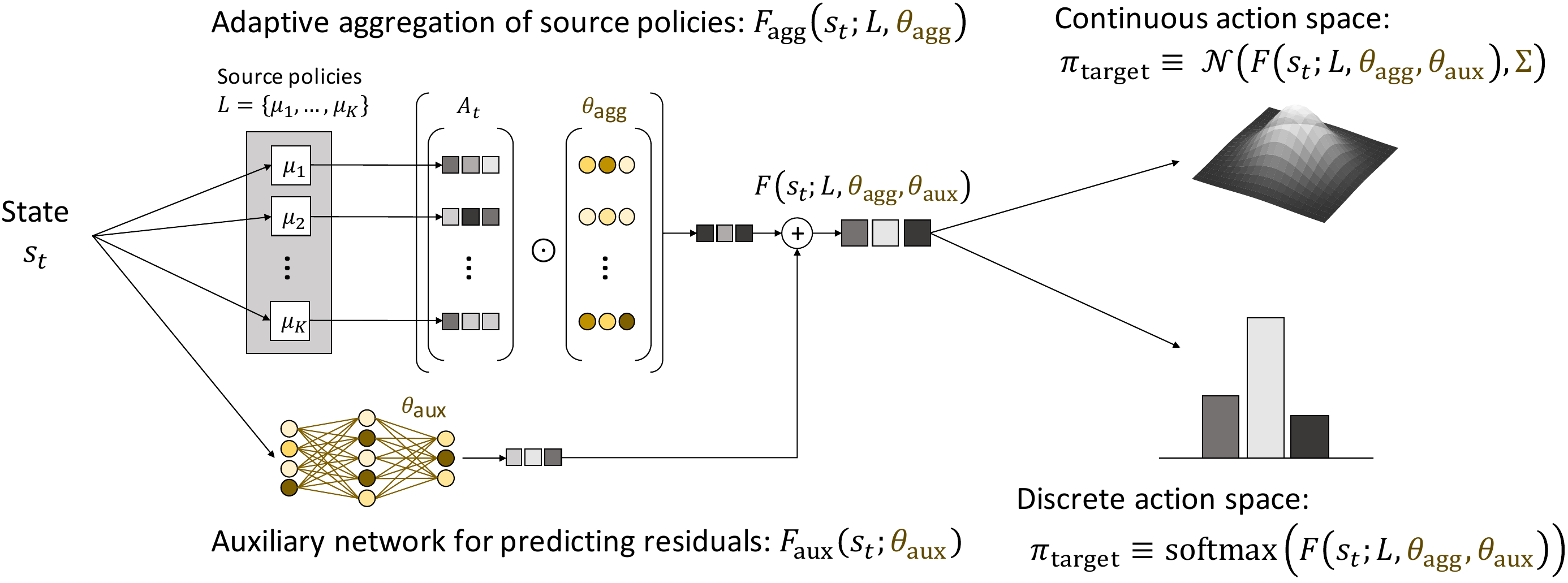}
\caption{Overview of MULTIPOLAR. We formulate a target policy $\pi\sub{target}$ with the sum of 1) the adaptive aggregation $F\sub{agg}$ of deterministic actions from source policies $L$ and 2) the auxiliary network $F\sub{aux}$ for predicting residuals around $F\sub{agg}$.}
\label{fig:MULTIPOLAR}
\end{figure*}

As a solution to the problem, we propose a new transfer RL approach named \textbf{MULTI-source POLicy AggRegation (MULTIPOLAR)}. As shown in Figure~\ref{fig:MULTIPOLAR}, our key idea is twofold; 1) In a target policy, we adaptively aggregate the deterministic actions produced by a collection of source policies. By learning aggregation parameters to maximize the expected return at a target environment instance, we can better adapt the aggregated actions to unseen environmental dynamics of the target instance without knowing source environmental dynamics nor source policy performances. 2) We also train an auxiliary network that predicts a residual around the aggregated actions, which is crucial for ensuring the expressiveness of the target policy even when some source policies are not useful. As another notable advantage, the proposed MULTIPOLAR can be used for both continuous and discrete action spaces with few modifications while allowing a target policy to be trained in a principled fashion.

We evaluate MULTIPOLAR in a variety of environments ranging from classic control problems to challenging robotics simulations. Our experimental results demonstrate the significant improvement of sample efficiency with the proposed approach, compared to baselines learning from scratch, as well as leveraging a single source policy and A2T. We also conducted a detailed analysis of our approach and found it worked well even when some of the source policies performed poorly in their original environment instance. 

\textbf{Main contributions}: (1) a new transfer RL problem that leverages multiple source policies collected under diverse unknown environmental dynamics to train a target policy in another dynamics, and (2) MULTIPOLAR, a principled and effective solution verified in our extensive experiments.

\section{Preliminaries} 
\label{sec:prelim}

\paragraph{Reinforcement Learning.}
We formulate our problem under the standard RL framework~\cite{Sutton1998}, where an agent interacts with its environment modeled by a Markov decision process (MDP). An MDP is represented by the tuple $\mathcal{M} = (\rho_0, \gamma, \mathcal{S}, \mathcal{A}, R, T)$ where $\rho_0$ is the initial state distribution and $\gamma$ is a discount factor. At each timestep $t$, given the current state $s_t\in\mathcal{S}$, the agent executes an action $a_t\in\mathcal{A}$ based on its policy $\pi(a_t\mid s_t;\theta)$ parameterized by $\theta$. Importantly, in this work, we consider both cases of continuous and discrete for action space $\mathcal{A}$. The environment returns a reward $R(s_t, a_t)\in\R$ and transitions to the next state $s_{t+1}$ based on the state transition distribution $T(s_{t+1}\mid s_t, a_t)$. In this framework, RL aims to maximize the expected return with respect to the policy parameters $\theta$.

\paragraph{Environment Instances.}
Similar to some prior works on transfer RL~\cite{Song2016,pmlr-v80-tirinzoni18a,yu2018policy}, we consider $K$ instances of the same environment which differ only in their state transition dynamics. Namely, we model each environment instance by an indexed MDP: $\mathcal{M}_i= (\rho_0, \gamma, \mathcal{S}, \mathcal{A}, R, T_i)$ where no two state transition distributions $T_i, T_j; i\neq j$ are identical. Unlike the prior works, we assume that \emph{each $T_i$ is unknown when training a target policy}, \ie, agents cannot access the exact form of $T_i$ nor a collection of states sampled from $T_i$.

\paragraph{Source Policies.}
For each of the $K$ environment instances, we are given a deterministic source policy $\mu_i:\mathcal{S}\rightarrow\mathcal{A}$ that only maps states to actions. Each source policy $\mu_i$ can be either parameterized (\eg, learned by interacting with its environment instance $\mathcal{M}_i$) or non-parameterized (\eg, heuristically designed by humans). Either way, we assume that \emph{no prior knowledge about the source policies $\mu_i$'s is available for a target agent, such as their representations or original performances, except that they were acquired from a source environment instance $\mathcal{M}_i$ with an unknown $T_i$}. This is one of the assumptions that make our problem unique from others, such as policy reuse and option frameworks. 

\paragraph{Problem Statement.} Given the set of source policies $L=\{\mu_1,\dots,\mu_K\}$, our goal is to train a new target agent's policy $\pi\sub{target}(a_t\mid s_t;L, \theta)$ in a sample efficient fashion, where the target agent interacts with another environment instance $\mathcal{M}\sub{target}=(\rho_0, \mathcal{S}, \mathcal{A}, R, T\sub{target})$ and $T\sub{target}$ is not identical to the source $T_i\; (i=1\dots,K)$ due to their distinct dynamics.

\section{Multi-Source Policy Aggregation}
\label{sec:multipolar}

As shown in Figure~\ref{fig:MULTIPOLAR}, with the Multi-Source Policy Aggregation (MULTIPOLAR), we formulate a target policy $\pi\sub{target}$ using a) the adaptive aggregation of deterministic actions from the set of source policies $L$, and b) the auxiliary network predicting residuals around the aggregated actions. We first present our method for the continuous action space, and then extend it to the discrete space.

\paragraph{Adaptive Aggregation of Source Policies.}
Let us denote by $a_t^{(i)}=\mu_i(s_t)$ the action predicted deterministically by source policy $\mu_i$ given the current state $s_t$. For the continuous action space, $a^{(i)}_t\in\mathbb{R}^{D}$ is a $D$-dimensional real-valued vector representing $D$ actions performed jointly in each timestep. For the collection of source policies $L$, we derive the matrix of their deterministic actions: 
\begin{equation}
A_t=\left[(a^{(1)}_t)^\top,\dots,(a^{(K)}_t)^\top\right] \in \displaystyle \R^{K\times D}.
\end{equation}
The key idea of this work is to aggregate $A_t$ adaptively in an RL loop, \ie, to maximize the expected return. This adaptive aggregation gives us a ``baseline'' action that could introduce a strong inductive bias in the training of a target policy \emph{even without knowing each source environmental dynamics $T_i$}. As mentioned earlier, other recent methods that enable transfer between different dynamics, by contrast, require access to the source environmental dynamics or related physical parameters \cite{Chen2018,yu2018policy}.

More specifically, we define the adaptive aggregation function $F\sub{agg}: \mathcal{S}\rightarrow \mathcal{A}$ that produces the baseline action based on the current state $s_t$ as follows:
\begin{equation}
    F\sub{agg}(s_t; L, \theta\sub{agg}) = \frac{1}{K}\mathds{1}^K \left(\theta\sub{agg}\odot A_t\right),
    \label{eq:baseline}
\end{equation}
where $\theta\sub{agg}\in\mathbb{R}^{K\times D}$ is a matrix of trainable parameters, $\odot$ is the element-wise multiplication, and $\mathds{1}^K$ is the all-ones vector of length $K$. $\theta\sub{agg}$ is neither normalized nor regularized, and can scale each action of each policy independently. This means that unlike A2T~\cite{rajendran:attend} method, we do not merely adaptively interpolate action spaces, but more flexibly emphasize informative source actions while suppressing irrelevant ones. Furthermore, in contrast to A2T method that trains a new network from scratch to map states to aggregation weights, our approach directly estimates $\theta\sub{agg}$ in a state-independent fashion, as differences in environmental dynamics would affect optimal actions in the entire state space rather than a part of it.

\paragraph{Predicting Residuals around Aggregated Actions.}
Moreover, we learn auxiliary network $F\sub{aux}: \mathcal{S}\rightarrow\mathcal{A}$ jointly with $F\sub{agg}$, to predict residuals around the aggregated actions. $F\sub{aux}$ is used to improve the target policy training in two ways. 1) If the aggregated actions from $F\sub{agg}$ are already useful in the target environment instance, $F\sub{aux}$ will correct them for a higher expected return. 2) Otherwise, $F\sub{aux}$ learns the target task while leveraging $F\sub{agg}$ as a prior to have a guided exploration process. Any network could be used for $F\sub{aux}$ as long as it is parameterized and fully differentiable. Finally, the MULTIPOLAR function is formulated as:
\begin{equation}
\resizebox{.91\linewidth}{!}{$
    \displaystyle
    F(s_t;L, \theta\sub{agg}, \theta\sub{aux}) = F\sub{agg}(s_t; L, \theta\sub{agg}) + F\sub{aux}(s_t; \theta\sub{aux}),$}
    \label{eq:multipolar}
\end{equation}
where $\theta\sub{aux}$ denotes the set of trainable parameters of $F\sub{aux}$. Note that the idea of predicting residuals for a source policy has also been presented in \cite{Silver2018,Johannink2018}. The main difference here is that, while these works just \emph{add} raw action outputs provided from a \emph{single} hand-engineered source policy, we adaptively aggregate actions from multiple source policies to exploit them flexibly.

\paragraph{Target Policy.}
Target policy $\pi\sub{target}$ can be modeled by reparameterizing the MULTIPOLAR function as a Gaussian distribution, \ie, $\mathcal{N}(F(s_t;L, \theta\sub{agg}, \theta\sub{aux}), \Sigma)$, where $\Sigma$ is a covariance matrix estimated based on what the used RL algorithm requires. Since we regard $\mu_i\in L$ as fixed functions mapping states to actions, this Gaussian policy $\pi\sub{target}$ is differentiable with respect to $\theta\sub{agg}$ and $\theta\sub{aux}$, and hence could be trained with any RL algorithm that explicitly updates policy parameters. Unlike~\cite{Silver2018,Johannink2018}, we can formulate the target policy in a principled fashion for actions in a discrete space. Specifically, instead of a $D$-dimensional real-valued vector, here we have a $D$-dimensional one-hot vector $a^{(i)}_t \in \{0, 1\}^D,\;\sum_j (a^{(i)}_t)_j = 1$ as outputs of $\mu_i$, where $(a^{(i)}_t)_j = 1$ indicates that the $j$-th action is to be executed. Following Eqs.~(\ref{eq:baseline}) and (\ref{eq:multipolar}), the output of $F(s_t;L, \theta\sub{agg}, \theta\sub{aux})$ can be viewed as $D$-dimensional un-normalized action scores, from which we can sample a discrete action after normalizing it by the softmax function.

\section{Experimental Evaluation}

We aim to empirically demonstrate the sample efficiency of a target policy trained with MULTIPOLAR (denoted by ``MULTIPOLAR policy''). To complete the experiments in a reasonable amount of time, we set the number of source policies to be $K=4$ unless mentioned otherwise. Moreover, we investigate the factors that affect the performance of MULTIPOLAR. To ensure fair comparisons and reproducibility of experiments, we followed the guidelines introduced by \cite{Sheila} and \cite{DBLP:journals/ftml/Francois-LavetH18} for conducting and evaluating all of our experiments.

\subsection{Experimental Setup}
\label{subsec:expSetup}

\paragraph{Baseline Methods.} To show the benefits of leveraging source policies, we compared our MULTIPOLAR policy to the standard multi-layer perceptron (MLP) trained from scratch, which is typically used in RL literature~\cite{ppo,DBLP:journals/ftml/Francois-LavetH18}. We also used MULTIPOLAR with $K=1$, which is an extension of residual policy learning~\cite{Silver2018,Johannink2018} (denoted by ``RPL'') with adaptive residuals as well as the ability to deal with both continuous and discrete action spaces. As another baseline, we used A2T~\cite{rajendran:attend} approach with the modification of replacing the actions sampled from stochastic source policies with deterministic source actions. We also made A2T employable in continuous action spaces by formulating the target policy in the same way as MULTIPOLAR explained in Section \ref{sec:multipolar}. We stress here that the existing transfer RL or meta RL approaches that train a universal policy network agnostic to the environmental dynamics, such as~\cite{Frans2017,Chen2018}, cannot be used as a baseline since they require a policy to be trained on a distribution of environment instances, which is not possible in our problem setting. Also, other techniques using multiple source policies, such as policy reuse frameworks~\cite{fernandez2006probabilistic}, \cite{parisotto:actormimic}, and \cite{yu2018policy}, are not applicable because they require source policies to be collected under the target environmental dynamics or to be trained simultaneously under known source dynamics.

\paragraph{Environments.} To show the general effectiveness of the MULTIPOLAR policy, we conducted comparative evaluations of MULTIPOLAR on the following six OpenAI Gym environments: Roboschool Hopper, Roboschool Ant, Roboschool InvertedPendulumSwingUp, Acrobot, CartPole, and LunarLander. We chose these six environments because 1) the parameterization of their dynamics and kinematics is flexible enough, 2) they cover discrete action space (Acrobot and CartPole) as well as continuous action space, and 3) they are samples of three distinct categories of OpenAI Gym environments, namely Box2d, Classic Control, and Roboschool. We used the Roboschool implementation of Hopper, Ant, and InvertedPendulumSwingup since they are based on an open-source engine, which makes it possible for every researcher to reproduce our experiments.

\paragraph{Experimental Procedure.}
For each of the six environments, we first created 100 environment instances by randomly sampling the dynamics and kinematics parameters from a specific range. For example, these parameters in the Hopper environment were link lengths, damping, friction, armature, and link mass with the sampling range defined similar to  \cite{Chen2018}. Details of sampling ranges for dynamics and kinematics parameters of all six environments are presented in Tables~\ref{tab:cart}, \ref{tab:acr}, \ref{tab:lunar}, \ref{tab:hopp}, \ref{tab:ant}, and \ref{tab:inv}. Note that we defined the sampling ranges for each environment such that the resulting environment instances are significantly different in dynamics. For this reason, a plain use of a source policy usually performed poorly in the target instance. Also, these parameters were used only for simulating environment instances and were not available when training a target policy. Then, for each environment instance, we trained an MLP policy that was used in two ways: a) the baseline MLP policy for each environment instance, and b) one of the 100 members of the source policy candidate pool from which we sample $K$ of them to train MULTIPOLAR policies as well as A2T policies, and one of them to train RPL policies\footnote{Although we used trained MLPs as source policies for reducing experiment times, any type of policies including hand-engineered ones could be used for MULTIPOLAR in principle.}. Specifically, for each environment instance, we trained three sets of MULTIPOLAR, A2T, and RPL policies each with distinct source policy sets selected randomly from the candidate pool. The learning procedure explained above was done three times with fixed different random seeds to reduce variance in results due to stochasticity. As a result, for each of the six environments, we had 100 environment instances $\times$ 3 random seeds $=$ 300 experiments for MLP and 100 environment instances $\times$ 3 choices of source policies $\times$ 3 random seeds $=$ 900 experiments for RPL, A2T, and MULTIPOLAR. The aim of this large number of experiments is to obtain correct insights into the distribution of performances~\cite{Sheila}. Due to the large number of experiments for all the environments, our detailed analysis and ablation study of MULTIPOLAR components were conducted only in Hopper environment, as its sophisticated second-order dynamics plays a crucial role in agent performance~\cite{Chen2018}.

\begin{table}[t]
\centering
\begin{minipage}{0.49\linewidth}
\centering
\scalebox{0.85}{
\begin{tabular}{@{}*{1}l *{1}l@{}}
\toprule
 \multicolumn{2}{c}{\textbf{Kinematics}} \\
\midrule
Links & Length Range\\
\midrule
Pole & {[0.1, 3]} \\
\\
\\
\\
\bottomrule
\end{tabular}
}
\end{minipage}
\begin{minipage}{0.49\linewidth}
\centering
\scalebox{0.85}{
\begin{tabular}{@{}*{1}l *{1}l@{}}
\toprule
\multicolumn{2}{c}{\textbf{Dynamics}} \\
\midrule
Factors & Value Range\\
\midrule
Force & {[6, 13]} \\
Gravity & {[-14, -6]} \\
Poll mass & {[0.1, 3]} \\
Cart mass & {[0.3, 4]} \\
\bottomrule
\end{tabular}
}
\end{minipage}
\caption{Sampling range for CartPole environment parameters.}
\label{tab:cart}
\end{table}

\begin{table}[t]
\centering
\begin{minipage}{0.47\linewidth}
\centering
\scalebox{0.85}{
\begin{tabular}{@{}*{1}l *{1}l@{}}
\toprule
 \multicolumn{2}{c}{\textbf{Kinematics}} \\
\midrule
Links & Length Range\\
\midrule
Link 1 & {[0.3, 1.3]} \\
Link 2 & {[0.3, 1.3]} \\
\\
\\
\bottomrule
\end{tabular}
}
\end{minipage}
\begin{minipage}{0.52\linewidth}
\centering
\scalebox{0.85}{
\begin{tabular}{@{}*{1}l *{1}l@{}}
\toprule
\multicolumn{2}{c}{\textbf{Dynamics}} \\
\midrule
Factors & Value Range\\
\midrule
Mass & {[0.5, 1.5]} \\
Center mass & {[0.05, 0.95]} \\
& $\times$ default length\\
Inertia moments & {[0.25, 1.5]} \\
\bottomrule
\end{tabular}
}
\end{minipage}
\caption{Sampling ranges for Acrobot environment parameters.}
\label{tab:acr}
\end{table}

\begin{table}[t]
\centering
\begin{minipage}{0.49\linewidth}
\centering
\scalebox{0.85}{
\begin{tabular}{@{}*{1}l *{1}l@{}}
\toprule
 \multicolumn{2}{c}{\textbf{Kinematics}} \\
\midrule
Links & Length Range\\
\midrule
Side engine & \multirow{2}{*}{[10, 20]} \\
height & \\
\\
\\
\\
\bottomrule
\end{tabular}
}
\end{minipage}
\begin{minipage}{0.49\linewidth}
\centering
\scalebox{0.85}{
\begin{tabular}{@{}*{1}l *{1}l@{}}
\toprule
\multicolumn{2}{c}{\textbf{Dynamics}} \\
\midrule
Factors & Value Range\\
\midrule
Scale & {[25, 50]} \\
Initial Random & {[500, 1500]} \\
Main engine power & {[10, 40]} \\
Side engine power & {[0.5, 2]} \\
Side engine away & {[8, 18]} \\
\bottomrule
\end{tabular}
}
\end{minipage}
\caption{Sampling ranges for LunarLander environment parameters.}
\label{tab:lunar}
\end{table}

\begin{table}[t]
\centering
\begin{minipage}{0.49\linewidth}
\centering
\scalebox{0.85}{
\begin{tabular}{@{}*{1}l *{1}l@{}}
\toprule
 \multicolumn{2}{c}{\textbf{Kinematics}} \\
\midrule
Links & Length Range\\
\midrule
Leg & {[0.35, 0.65]} \\
Foot & {[0.29, 0.49]} \\
Thigh & {[0.35, 0.55]} \\
Torso & {[0.3, 0.5]} \\
\\
\bottomrule
\end{tabular}
}
\end{minipage}
\begin{minipage}{0.49\linewidth}
\centering
\scalebox{0.85}{
\begin{tabular}{@{}*{1}l *{1}l@{}}
\toprule
\multicolumn{2}{c}{\textbf{Dynamics}} \\
\midrule
Factors & Value Range\\
\midrule
Damping & {[0.5, 4]}  \\
Friction & {[0.5, 2]} \\
Armature & {[0.5, 2]} \\
Links mass & {[0.7, 1.1]} \\
& $\times$ default mass\\
\bottomrule
\end{tabular}
}
\end{minipage}
\caption{Sampling ranges for Hopper environment parameters.}
\label{tab:hopp}
\end{table}

\begin{table}[t]
\centering
\begin{minipage}{0.49\linewidth}
\centering
\scalebox{0.85}{
\begin{tabular}{@{}*{1}l *{1}l@{}}
\toprule
 \multicolumn{2}{c}{\textbf{Kinematics}} \\
\midrule
Links & Length Range\\
\midrule
Legs & {[0.4, 1.4]}  \\
& $\times$ default length\\
\\
\\
\\
\bottomrule
\end{tabular}
}
\end{minipage}
\begin{minipage}{0.49\linewidth}
\centering
\scalebox{0.85}{
\begin{tabular}{@{}*{1}l *{1}l@{}}
\toprule
\multicolumn{2}{c}{\textbf{Dynamics}} \\
\midrule
Factors & Value Range\\
\midrule
Damping & {[0.1, 5]}  \\
Friction & {[0.4, 2.5]} \\
Armature & {[0.25, 3]} \\
Links mass & {[0.7, 1.1]} \\
& $\times$ default mass\\
\bottomrule
\end{tabular}
}
\end{minipage}
\caption{Sampling range for Ant environment parameters.}
\label{tab:ant}
\end{table}

\begin{table}[t]
\centering
\begin{minipage}{0.49\linewidth}
\centering
\scalebox{0.85}{
\begin{tabular}{@{}*{1}l *{1}l@{}}
\toprule
 \multicolumn{2}{c}{\textbf{Kinematics}} \\
\midrule
Links & Length Range\\
\midrule
Pole & {[0.2, 2]} \\
\\
\\
\\
\\
\\
\bottomrule
\end{tabular}
}
\end{minipage}
\begin{minipage}{0.49\linewidth}
\centering
\scalebox{0.85}{
\begin{tabular}{@{}*{1}l *{1}l@{}}
\toprule
\multicolumn{2}{c}{\textbf{Dynamics}} \\
\midrule
Factors & Value Range\\
\midrule
Damping & {[0.1, 5]}  \\
Friction & {[0.5, 2]} \\
Armature & {[0.5, 3]} \\
Gravity & {[-11, -7]} \\
Links mass & {[0.4, 3]} \\
& $\times$ default mass\\
\bottomrule
\end{tabular}
}
\end{minipage}
\caption{Sampling ranges for InvertedPendulumSwingup environment parameters.}
\label{tab:inv}
\end{table}

\paragraph{Implementation Details.}
All the experiments were done using the Stable Baselines~\cite{stable-baselines} implementation of learning algorithms as well as its default hyperparameters and MLP network architecture for each environment. Based on the performance of learning algorithms reported in \cite{stable-baselines}, all the policies were trained with Soft Actor-Critic~\cite{SAC} in the LunarLander environment and with Proximal Policy Optimization~\cite{ppo} in the rest of the environments. For fair comparisons, in all experiments, auxiliary network $F\sub{aux}$ had an identical architecture to that of the MLP. Therefore, the only difference between MLP and MULTIPOLAR was the aggregation part $F\sub{agg}$, which made it possible to evaluate the contribution of transfer learning based on adaptive aggregation of source policies. In the same way as~\cite{rajendran:attend}, to implement the attention network branch of A2T, we used the same MLP architecture except for the last layer, which is activated with a softmax function. In all the experiments, source policies are the same for A2T and MULTIPOLAR to ensure an unbiased evaluation. We avoided any random seed optimization since it has been shown to alter the policies' performance~\cite{Sheila}. As done by \cite{stable-baselines}, to have a successful training, we normalized rewards and input observations using their running average and standard deviation for all the environments except CartPole and LunarLander. Furthur, in all of the experiments, $\theta\sub{agg}$ is initialized to be the all-ones matrix. To run our experiments in parallel, we used GNU Parallel tool~\cite{tange_ole_2018_1146014}. Finally, all the hyperparameters used for experiments on each environment can be found in our codebase, which is available on the project website: \url{https://omron-sinicx.github.io/multipolar/}.

\paragraph{Evaluation Metric.} Following the guidelines of \cite{Sheila}, to measure sampling efficiency of training policies, \ie, how quick the training progresses, we used the average episodic reward over training samples. Also, to ensure that higher average episodic reward is representative of better performance and to estimate the variation of it, we used the sample bootstrap method to estimate statistically relevant 95\% confidence bounds of the results of our experiments. Across all the experiments, we used 10K bootstrap iterations and the pivotal method\footnote{We used the Facebook Boostrapped implementation: \url{github.com/facebookincubator/bootstrapped}.}.

\begin{table*}[tp]
\centering
\begin{minipage}{0.49\linewidth}
\scalebox{0.95}{
\begin{tabular}{@{} l SS @{}} 
\toprule

\multirow{2}{*}{Methods}    &  \multicolumn{2}{c @{}}{CartPole}\\ 
\cmidrule(l){2-3}
     & {50K} & {100K} \\
\midrule
MLP     & {229 (220,237)}  &  {291 (282,300)} \\
RPL     & {238 (231,245)} & {289 (282,296)} \\
A2T (K=4) & {230 (224,235)} & {281 (275,287)} \\
MULTIPOLAR (K=4)  & {\textbf{252 (245,260)}}  & {\textbf{299 (292,306)}} \\
\midrule\midrule

  &  \multicolumn{2}{c @{}}{Acrobot}\\ 
\cmidrule(l){2-3}
    & {100K} & {200K} \\
\midrule
MLP        & {-164 (-172,-156)} & {-111 (-117,-106)} \\
RPL & {-120 (-124,-116)} & {-98 (-101,-95)} \\
A2T (K=4) & {-201 (-208,-195)} & {-120 (-123, -116)} \\
MULTIPOLAR (K=4)  & {\textbf{-117 (-121,-113)}} & {\textbf{-96 (-99,-93)}} \\
\midrule\midrule

 &  \multicolumn{2}{c @{}}{LunarLander}\\ 
\cmidrule(l){2-3}
    & {250K} & {500K} \\
\midrule
MLP        &  {112 (104,121)} &  {216 (210,221)} \\
RPL &   {178 (174,182)} & {\textbf{246 (243,248)}} \\
A2T (K=4) & {128 (123,133)} & {226 (223,229)} \\
MULTIPOLAR (K=4)  & {\textbf{181 (177,185)}} & {\textbf{246 (244,249)}} \\
\bottomrule
\end{tabular}
}
\end{minipage}
\begin{minipage}{0.49\linewidth}
\scalebox{0.95}{
\begin{tabular}{@{} l SS @{}} 
\toprule
\multirow{2}{*}{Methods}   &  \multicolumn{2}{c @{}}{Roboschool Hopper}\\ 
\cmidrule(l){2-3}
    & {1M} & {2M} \\
\midrule
MLP         & {43 (42,45)} &  {92 (88,96)} \\
RPL & {75 (70,79)} & {152 (142,160)} \\
A2T (K=4) & {58 (56,59)} & {126 (122,129)} \\
MULTIPOLAR (K=4)   & {\textbf{138 (132,143)}} & {\textbf{283 (273,292)}} \\
\midrule\midrule

  &  \multicolumn{2}{c @{}}{Roboschool Ant}\\ 
\cmidrule(l){2-3}
    & {1M}  & {2M} \\
\midrule
MLP         & {1088 (1030,1146)} &  {1500 (1430,1572)} \\
RPL & {1120 (1088,1152)} & {1432 (1391,1473)} \\
A2T (K=4) & {1025 (990,1059)} & {1361 (1320,1402)} \\
MULTIPOLAR (K=4)  & {\textbf{1397 (1361,1432)}} & {\textbf{1744 (1705,1783)}} \\
\midrule\midrule

  &  \multicolumn{2}{c @{}}{Roboschool InvertedPendulumSwingup}\\ 
\cmidrule(l){2-3}
    &{1M} & {2M} \\
\midrule
MLP         & {267 (260,273)} &  {409 (401,417)} \\
RPL & {195 (192,198)} & {322 (317,326)} \\
A2T (K=4) & {310 (305,315)} & {458 (452,464)} \\
MULTIPOLAR (K=4)  & {\textbf{476 (456,495)}} & {\textbf{588 (571,605)}} \\
\bottomrule
\end{tabular}
}
\end{minipage}
\caption{Bootstrap mean and 95\% confidence bounds of average episodic rewards over various training samples across six environments.} 
\label{tab:results}
\end{table*}

\begin{figure*}[t]
\centering
\minipage{0.33\linewidth}%
  \includegraphics[width=\linewidth]{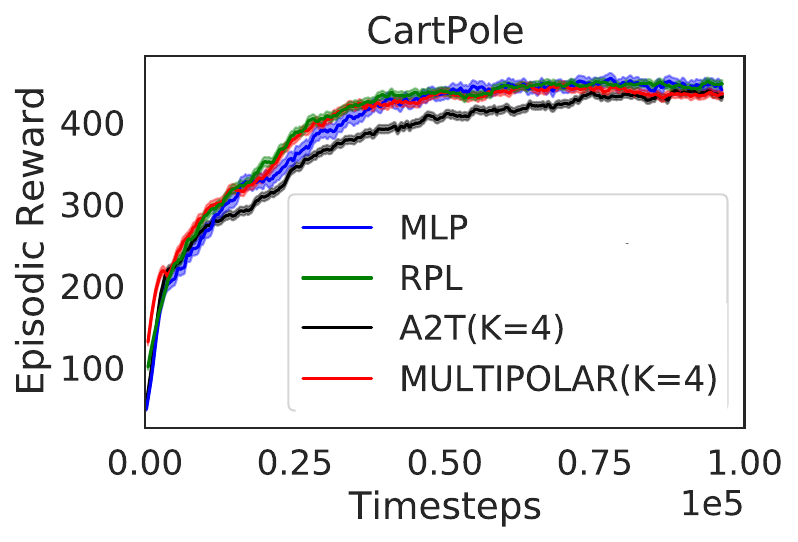}
\endminipage
\minipage{0.33\linewidth}%
  \includegraphics[width=\linewidth]{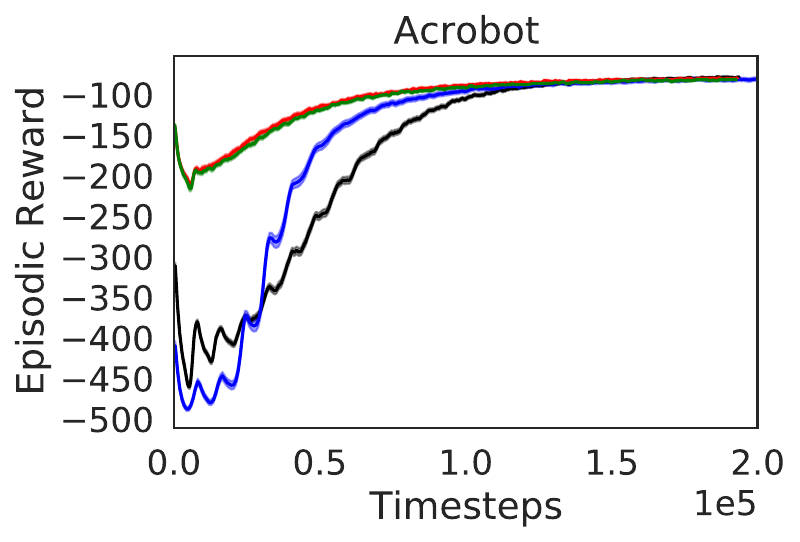}
\endminipage
\minipage{0.33\linewidth}%
  \includegraphics[width=\linewidth]{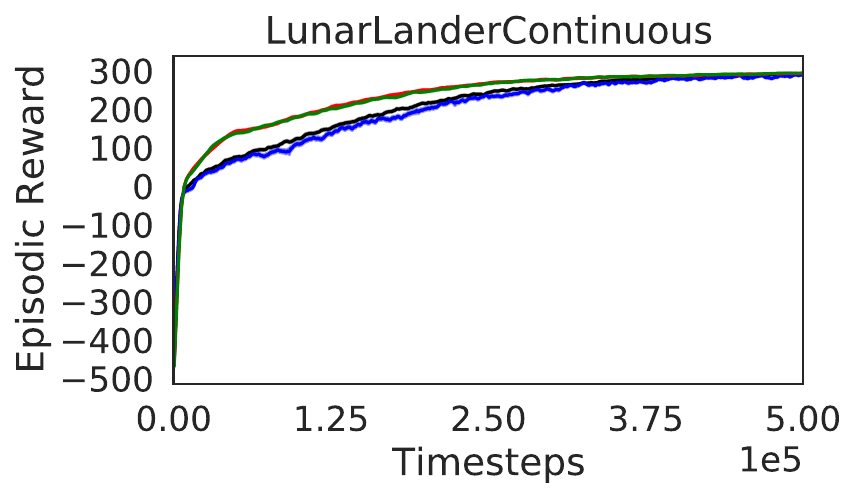}
\endminipage
\\
\minipage{0.33\linewidth}
  \includegraphics[width=\linewidth]{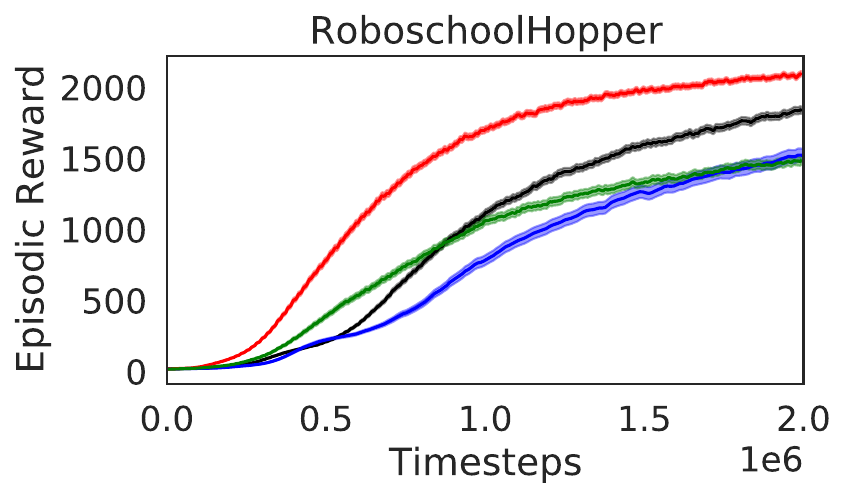}
\endminipage
\minipage{0.33\linewidth}
  \includegraphics[width=\linewidth]{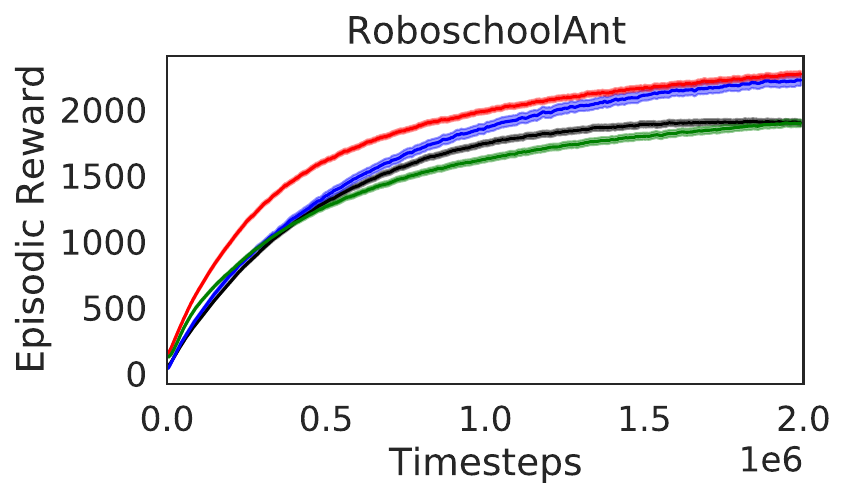}
\endminipage
\minipage{0.33\linewidth}%
  \includegraphics[width=\linewidth]{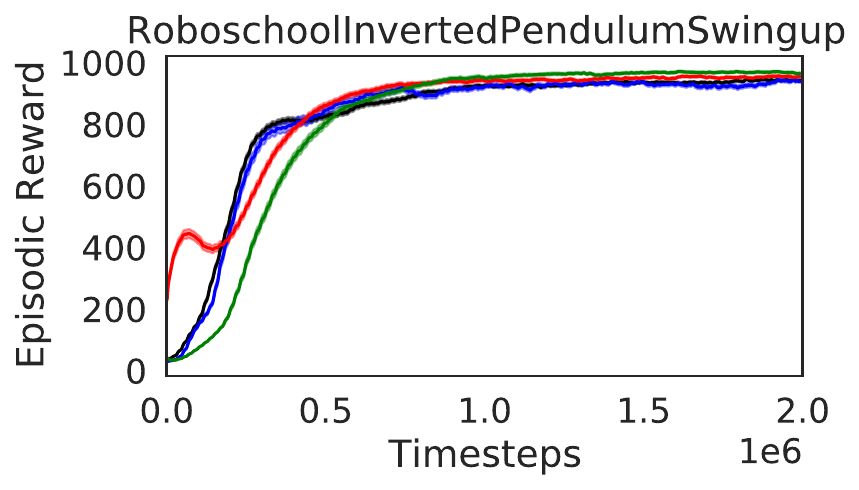}
\endminipage

\caption{Average learning curves of MLP, RPL, A2T ($K=4$), and MULTIPOLAR ($K=4$) over all the experiments for each environment. The shaded area represents 1 standard error.}
\label{fig:learning}
\end{figure*}

\subsection{Results}
\label{subsec:results}

\paragraph{Sample Efficiency.} Figure~\ref{fig:learning} and Table~\ref{tab:results} clearly show that on average, in all the environments, MULTIPOLAR outperformed baseline policies in terms of sample efficiency and sometimes the final episodic reward\footnote{Episodic rewards in Figure~\ref{fig:learning} are averaged over 3 random seeds and 3 random source policy sets on 100 environment instances. Table~\ref{tab:results} reports the mean of this average over training samples.}. For example, in Hopper over 2M training samples, MULTIPOLAR with $K=4$ achieved a mean of average episodic reward about three times higher than MLP (\ie, training from scratch) and about twice higher than A2T and RPL. It is also noteworthy that MULTIPOLAR had always on par or better performance than RPL, which indicates the effectiveness of leveraging multiple source policies. Table~\ref{tab:results} further suggests that although A2T is considerably more data efficient than learning from scratch in 3 out of 6 environments, it is substantially outperformed by MULTIPOLAR in all of the environments. We believe that the A2T performance limitation stems from 1) requiring training an extra network branch (attention) with the same size as the auxiliary network which almost doubles the number of parameters to be trained and 2) soft-attention mechanism that assigns a single weight for each policy resulting in merely \emph{interpolation} of source policies and the auxiliary network, unlike our approach that flexibly aggregates each action of each source policy.

\paragraph{Ablation Study.} 
To demonstrate the importance of each component of MULTIPOLAR, we evaluated the following degraded versions: (1) \emph{$\theta\sub{agg}$ fixed to 1}, which just averages the deterministic actions from the source policies without adaptive weights (similar to the residual policy learning methods that use raw action outputs of a source policy), and (2) \emph{$F\sub{aux}$ learned independent of $s_t$}, which replaces the state-dependent MLP with an adaptive ``placeholder'' parameter vector making actions a linear combination of source policy outputs. As shown in Table~\ref{tab:ablation}, the full version of MULTIPOLAR significantly outperformed the degraded ones, suggesting that adaptive aggregation and predicting residuals are both critical.

\begin{table}[t]
\centering
\scalebox{0.82}{
\begin{tabular}{@{}l c c@{}}
\toprule
MULTIPOLAR (K=4) & {1M} & {2M} \\
\midrule
Full version  & \textbf{{138 (132,143)}} &  \textbf{{283 (273,292)}} \\
{ $\theta\sub{agg}$ fixed to 1} & {118 (111,126)} &  {237 (222,250)} \\
$F\sub{aux}$ learned independent of $s_t$ & {101 (95,108)} &  {187 (175,200)} \\
\bottomrule
\end{tabular}
}
\caption{Ablation study in Hopper.}
\label{tab:ablation}
\end{table}

\begin{figure*}[t!]
\centering
\minipage{0.30\linewidth}%
  \includegraphics[width=\linewidth]{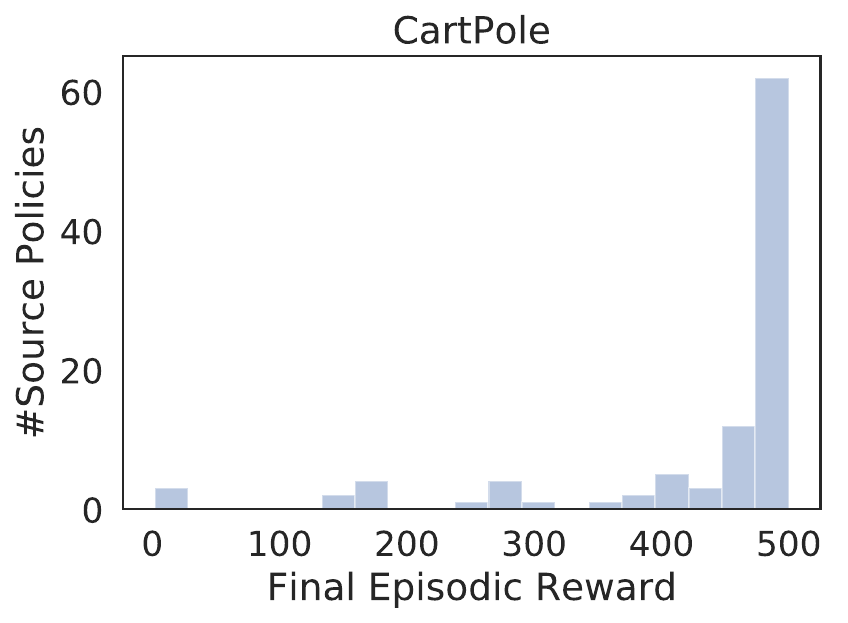}
\endminipage
\minipage{0.30\linewidth}%
  \includegraphics[width=\linewidth]{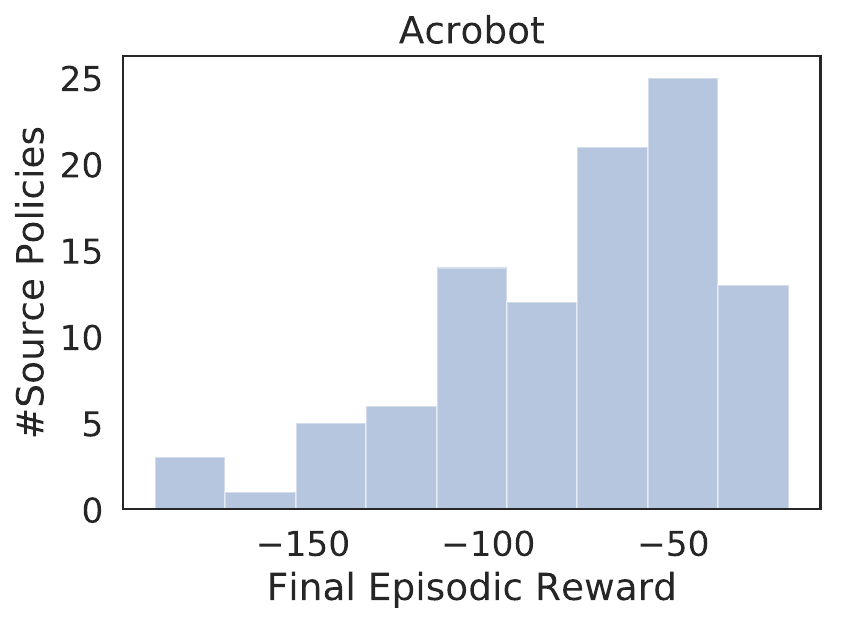}
\endminipage
\minipage{0.30\linewidth}%
  \includegraphics[width=\linewidth]{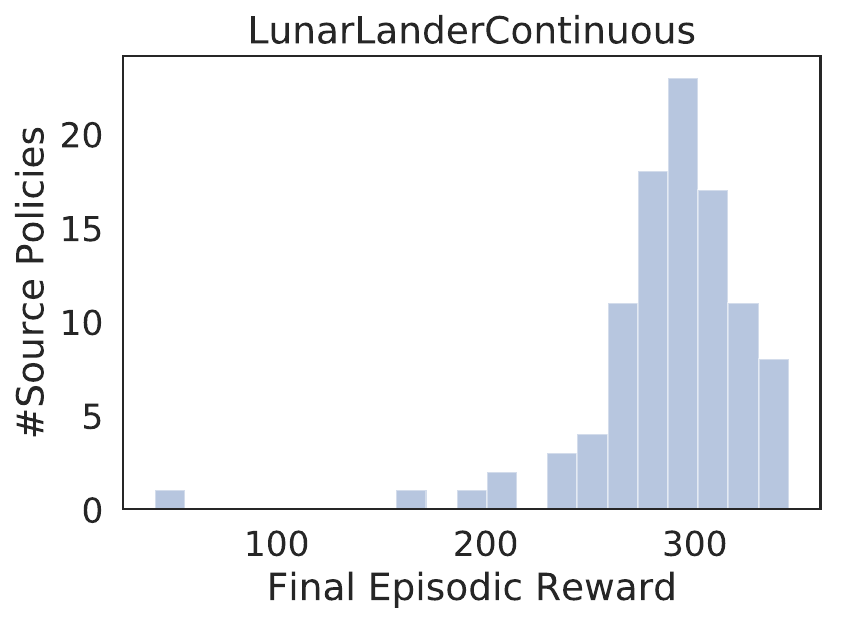}
\endminipage
\\
\minipage{0.30\linewidth}
  \includegraphics[width=\linewidth]{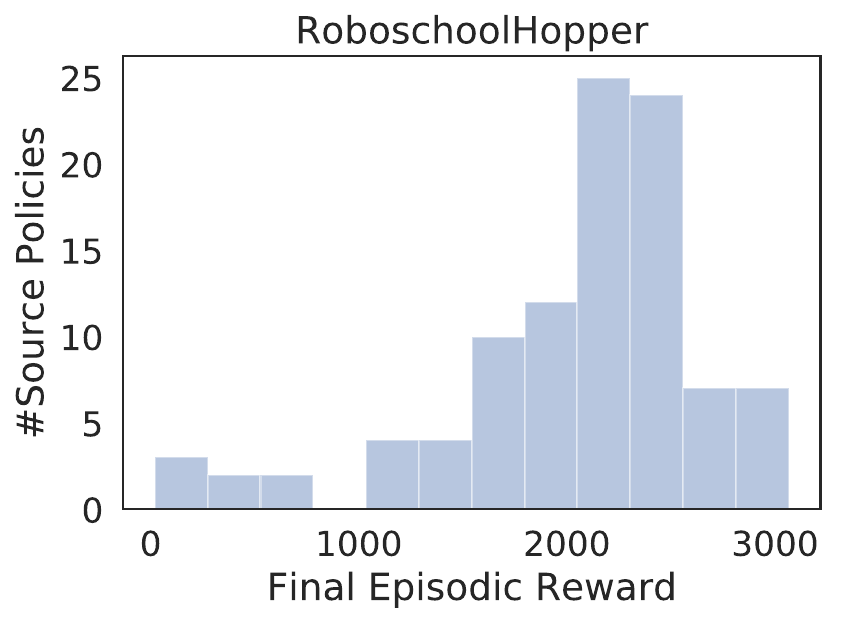}
\endminipage
\minipage{0.30\linewidth}
  \includegraphics[width=\linewidth]{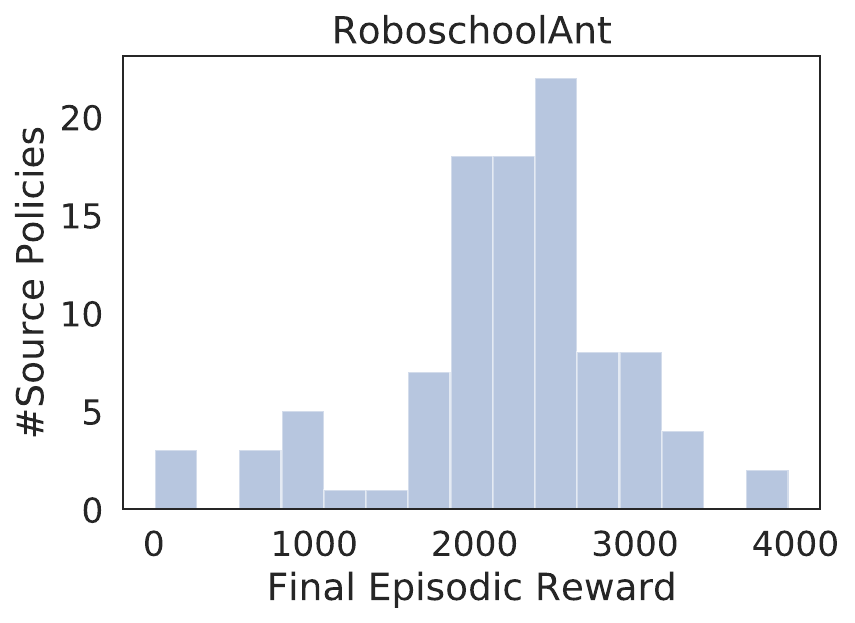}
\endminipage
\minipage{0.30\linewidth}%
  \includegraphics[width=\linewidth]{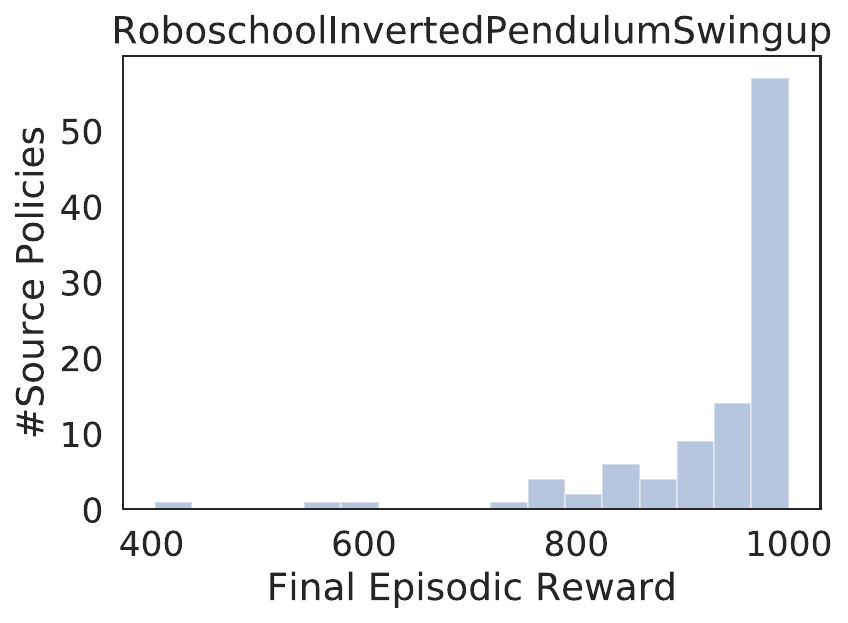}
\endminipage

\caption{Histogram of final episodic rewards obtained by source policies in their original environment instances.}
\label{fig:hist-all}
\end{figure*}

\begin{figure}[t]
  \begin{center}
    \includegraphics[width=\linewidth]{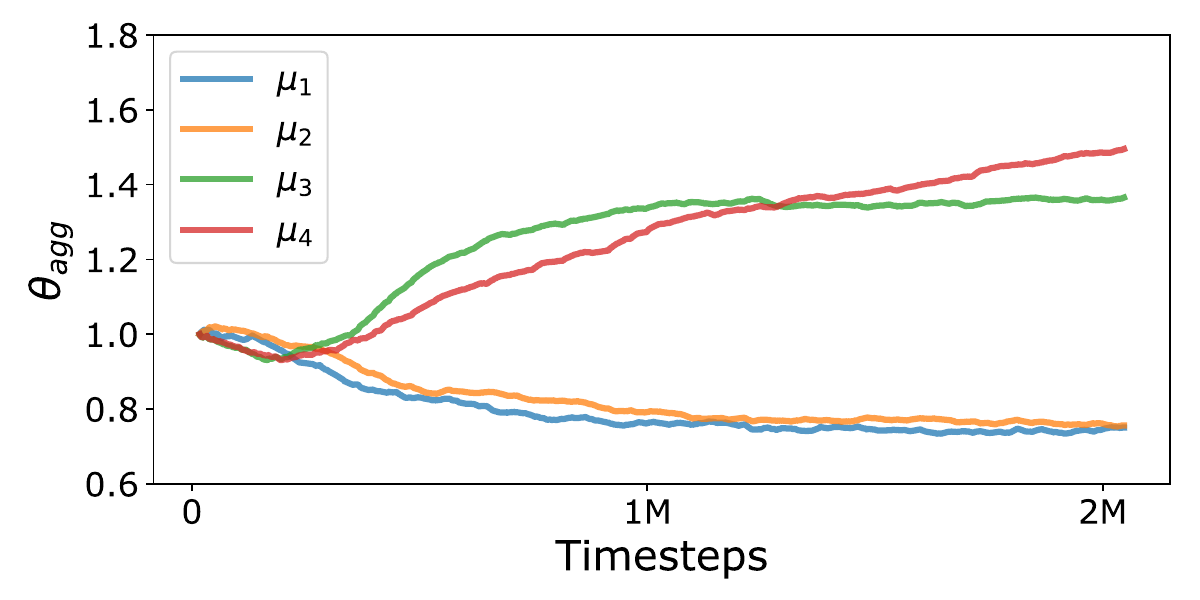}
    \caption{Aggregation parameters $\theta\sub{agg}$ averaged over the three actions of Hopper environment during the training, where source policies $\mu_1, \mu_2$ are low-performing and $\mu_3, \mu_4$ are high-performing.}
    \label{fig:weights}
  \end{center}
\end{figure}

\paragraph{Effect of Source Policy Performances.}
Figure~\ref{fig:hist-all} shows the histograms of final episodic rewards (average rewards of the last 100 training episodes) for the source policy candidates obtained in their own original environment instances. As shown in the figure, the source policies were diverse in terms of the performance. In this setup, we investigate the effect of source policies performances on MULTIPOLAR sample efficiency. We created two separate pools of source policies, where one contained only high-performing and the other only low-performing ones\footnote{Here, policies with final episodic reward over 2K are regarded as high-performing and below 1K are as low-performing.}. Table~\ref{tab:optimality} summarizes the results of sampling source policies from these pools (4 high, 2 high \& 2 low, and 4 low performances) and compares them to the original MULTIPOLAR (shown as `Random') also reported in Table~\ref{tab:results}. Not surprisingly, MULTIPOLAR performed the best when all the source policies were sampled from the high-performance pool. However, we emphasize that such high-quality policies are not always available in practice, due to the variability of how they are learned or hand-crafted under their own environment instance. Interestingly, by comparing the reported results in Table 2 (MLP) and Table 4 (4 low performance), we can observe that even in the worst case scenario of having only low performing source policies, the sample efficiency of MULTIPOLAR is on par with that of learning from scratch. This suggests that MULTIPOLAR avoids negative transfer, which occurs when transfer degrades the learning efficiency instead of helping it. Further, Figure~\ref{fig:weights} shows an example where MULTIPOLAR successfully learned to suppress the useless low-performing sources to maximize the expected return in a target task, indicating the mechanism of source policy rejection.

\begin{table}[t]
\centering
\scalebox{.9}{
\begin{tabular}{@{}l c c@{}}
\toprule
MULTIPOLAR (K=4) & {1M}  & {2M} \\
\midrule
Random & {138 (132,143)}  &  {283 (273,292)} \\
4 high performance  & \textbf{{214 (208,220)}}  &  \textbf{{420 (409,430)}} \\
2 high \& 2 low performance & {98 (94,102)} &  {208 (200,215)} \\
4 low performance  & {45 (44,47)} &  {92 (88,95)} \\
\bottomrule
\end{tabular}
}
\caption{Results with different source policy sampling in Hopper.}
\label{tab:optimality}
\end{table}

\begin{table}[t]
\centering
\small
\begin{tabular}{@{}l c c@{}}
\toprule
MULTIPOLAR & {1M}  & {2M} \\
\midrule
K=4 & {138 (132,143)}  &  {283 (273,292)} \\
K=8 & {160 (154,167)}  &  {323 (312,335)} \\
K=16  & \textbf{{177 (172,182)}} & \textbf{{357 (348,367)}} \\
\bottomrule
\end{tabular}
\caption{Results with different number of source policies \\in Hopper.}
\label{tab:number}
\end{table}

\paragraph{Effect of Number of Source Policies.} 
Finally, we show how the number of source policies contributes to MULTIPOLAR's sample efficiency in Table~\ref{tab:number}. Specifically, we trained MULTIPOLAR policies up to $K=16$ to study how the mean of average episodic rewards changes. The monotonic performance improvement over $K$ (for $K\leq 16$), is achieved at the cost of increased training and inference time. In practice, we suggest balancing this speed-performance trade-off by using as many source policies as possible before reaching the inference time limit required by the application.

\section{Discussion and Related Work}  
\label{sec:related_work}
In this section, we highlight how our work is different from the existing approaches and also discuss the future directions.

\paragraph{Transfer between Different Dynamics.}
Our work is broadly categorized as an instance of transfer RL between different environmental dynamics, in which a policy for a target task is trained using information collected from source tasks. Much related work requires training samples collected from source tasks, which are then used for measuring the similarity between source and target environment instances~\cite{Lazaric2008,pmlr-v80-tirinzoni18a} or conditioning a target policy to predict actions~\cite{Chen2018}. Alternative means to quantify the similarity is to use a full specification of MDPs~\cite{Song2016,Wang2019} or environmental dynamics~\cite{yu2018policy}.  The work of \cite{parisotto:actormimic} presented a multi-task learning framework to enable transfer between different games. However, it requires a target policy to be pre-trained with a variety of source tasks. In contrast, the proposed MULTIPOLAR allows the knowledge transfer only through the policies acquired from source environment instances with diverse unknown dynamics, which is beneficial when source and target environments are not always connected to exchange information about their dynamics and training samples.

\paragraph{Leveraging Multiple Policies.}
The idea of utilizing multiple source policies can be found in the literature of policy reuse frameworks~\cite{fernandez2006probabilistic,rosman2016bayesian,li2018optimal,zheng2018deep,Li2018}. The basic motivation behind these works is to provide ``nearly-optimal solutions''~\cite{rosman2016bayesian} for short-duration tasks by reusing one of the source policies, where each source would perform well on environment instances with different rewards (\eg, different goals in maze tasks). In our problem setting, where environmental dynamics behind each source policy are different, reusing a single policy is not the right approach as described in \cite{Chen2018}. Even with leveraging multiple source policies by a convex combination where the weights are predicted from an additional network learned from scratch~\cite{rajendran:attend}, the performance is still limited without effective and flexible source policy aggregation, as demonstrated in our experiments.
Another relevant idea is hierarchical RL~\cite{kulkarni2016hierarchical,osa2019hierarchical} that involves a hierarchy of policies (or action-value functions) to enable temporal abstraction. In particular, option frameworks~\cite{sutton1999between,Bacon2016,mankowitz2018learning} make use of a collection of policies as a part of ``options''. However, they assumed all the policies in the hierarchy to be learned in a single environment instance. Another relevant work along this line of research is \cite{Frans2017}, which meta-learns a hierarchy of multiple sub-policies by training a master policy over the distribution of tasks. Nevertheless, hierarchical RL approaches are not useful for leveraging multiple source policies each acquired under diverse environmental dynamics.

\paragraph{Learning Residuals in RL.} Some recent works adopt residual learning to mitigate the limited performance of hand-engineered policies \cite{Silver2018,Johannink2018}. We are interested in a more extended scenario where various source policies with unknown performances are provided instead of a single sub-optimal policy. 
Also, these approaches focus only on robotic tasks in the continuous action space, while our approach could work on both of continuous and discrete action spaces in a broad range of environments.

\paragraph{Future Directions.} In our experiments, we confirmed the effectiveness of our proposed MULTIPOLAR with up to 16 source policies. An important question is up to how many source policies will eventually saturate the performance gain of MULTIPOLAR? Another interesting direction is adapting MULTIPOLAR to combinatorial optimization problems~\cite{bello2016neural} (with heuristic approaches as source policies) as well as involving other types of environmental differences, such as dissimilar reward functions and state/action spaces.


\section{Conclusion} 
We presented a new problem setting of transfer RL which aims to train a policy efficiently using a collection of source policies acquired under diverse environmental dynamics. To this end, we proposed MULTIPOLAR that adaptively aggregates a set of actions provided by the source policies while learning a residual around the aggregated actions. This approach can be adopted for both continuous and discrete action spaces and is particularly advantageous when one does not have access to a distribution of source environment instances with diverse dynamics. We confirmed the high training sample efficiency of our approach on a variety of environments. Future work seeks to extend MULTIPOLAR to other challenging problems such as sim-to-real transfer~\cite{tan2018} and real-world robotics tasks.

\section*{Acknowledgments}
The authors would like to thank Robert Lee, Felix von Drigalski, Yoshihisa Ijiri, Tatsunori Taniai, and Daniel Plop, for the insightful discussions and helpful feedback on the manuscript.

\bibliographystyle{named}
\bibliography{iclr2020_conference}

\begin{thebibliography}{}

\bibitem[\protect\citeauthoryear{Bacon \bgroup \em et al.\egroup
  }{2017}]{Bacon2016}
Pierre-Luc Bacon, Jean Harb, and Doina Precup.
\newblock {The Option-Critic Architecture}.
\newblock In {\em AAAI}, 2017.

\bibitem[\protect\citeauthoryear{Bello \bgroup \em et al.\egroup
  }{2016}]{bello2016neural}
Irwan Bello, Hieu Pham, Quoc~V Le, Mohammad Norouzi, and Samy Bengio.
\newblock Neural combinatorial optimization with reinforcement learning.
\newblock {\em arXiv preprint arXiv:1611.09940}, 2016.

\bibitem[\protect\citeauthoryear{Chen \bgroup \em et al.\egroup
  }{2018}]{Chen2018}
Tao Chen, Adithyavairavan Murali, and Abhinav Gupta.
\newblock {Hardware Conditioned Policies for Multi-Robot Transfer Learning}.
\newblock In {\em NeurIPS}, 2018.

\bibitem[\protect\citeauthoryear{Clavera \bgroup \em et al.\egroup
  }{2019}]{clavera2018learning}
Ignasi Clavera, Anusha Nagabandi, Simin Liu, Ronald~S. Fearing, Pieter Abbeel,
  Sergey Levine, and Chelsea Finn.
\newblock {Learning to Adapt in Dynamic, Real-World Environments through
  Meta-Reinforcement Learning}.
\newblock In {\em ICLR}, 2019.

\bibitem[\protect\citeauthoryear{Devin \bgroup \em et al.\egroup
  }{2017}]{devin2017learning}
Coline Devin, Abhishek Gupta, Trevor Darrell, Pieter Abbeel, and Sergey Levine.
\newblock {Learning Modular Neural Network Policies for Multi-Task and
  Multi-Robot Transfer}.
\newblock In {\em ICRA}, 2017.

\bibitem[\protect\citeauthoryear{Fern{\'a}ndez and
  Veloso}{2006}]{fernandez2006probabilistic}
Fernando Fern{\'a}ndez and Manuela Veloso.
\newblock {Probabilistic Policy Reuse in a Reinforcement Learning Agent}.
\newblock In {\em AAMAS}, 2006.

\bibitem[\protect\citeauthoryear{Fran{\c{c}}ois{-}Lavet \bgroup \em et
  al.\egroup }{2018}]{DBLP:journals/ftml/Francois-LavetH18}
Vincent Fran{\c{c}}ois{-}Lavet, Peter Henderson, Riashat Islam, Marc~G.
  Bellemare, and Joelle Pineau.
\newblock {An Introduction to Deep Reinforcement Learning}.
\newblock {\em Foundations and Trends in Machine Learning}, 11(3-4):219--354,
  2018.

\bibitem[\protect\citeauthoryear{Frans \bgroup \em et al.\egroup
  }{2018}]{Frans2017}
Kevin Frans, Jonathan Ho, Xi~Chen, Pieter Abbeel, and John Schulman.
\newblock {Meta Learning Shared Hierarchies}.
\newblock In {\em ICLR}, 2018.

\bibitem[\protect\citeauthoryear{Haarnoja \bgroup \em et al.\egroup
  }{2018}]{SAC}
Tuomas Haarnoja, Aurick Zhou, Pieter Abbeel, and Sergey Levine.
\newblock {Soft Actor-Critic: Off-Policy Maximum Entropy Deep Reinforcement
  Learning with a Stochastic Actor}.
\newblock In {\em ICML}, 2018.

\bibitem[\protect\citeauthoryear{Henderson \bgroup \em et al.\egroup
  }{2018}]{Sheila}
Peter Henderson, Riashat Islam, Philip Bachman, Joelle Pineau, Doina Precup,
  and David Meger.
\newblock {Deep Reinforcement Learning That Matters}.
\newblock In {\em AAAI}, 2018.

\bibitem[\protect\citeauthoryear{Hill \bgroup \em et al.\egroup
  }{2018}]{stable-baselines}
Ashley Hill, Antonin Raffin, Maximilian Ernestus, Adam Gleave, Rene Traore,
  Prafulla Dhariwal, Christopher Hesse, Oleg Klimov, Alex Nichol, Matthias
  Plappert, Alec Radford, John Schulman, Szymon Sidor, and Yuhuai Wu.
\newblock {Stable Baselines}.
\newblock \url{https://github.com/hill-a/stable-baselines}, 2018.

\bibitem[\protect\citeauthoryear{Johannink \bgroup \em et al.\egroup
  }{2019}]{Johannink2018}
Tobias Johannink, Shikhar Bahl, Ashvin Nair, Jianlan Luo, Avinash Kumar,
  Matthias Loskyll, Juan~Aparicio Ojea, Eugen Solowjow, and Sergey Levine.
\newblock {Residual Reinforcement Learning for Robot Control}.
\newblock In {\em ICRA}, 2019.

\bibitem[\protect\citeauthoryear{Kulkarni \bgroup \em et al.\egroup
  }{2016}]{kulkarni2016hierarchical}
Tejas~D Kulkarni, Karthik Narasimhan, Ardavan Saeedi, and Josh Tenenbaum.
\newblock {Hierarchical Deep Reinforcement Learning: Integrating Temporal
  Abstraction and Intrinsic Motivation}.
\newblock In {\em NeurIPS}, 2016.

\bibitem[\protect\citeauthoryear{Lazaric \bgroup \em et al.\egroup
  }{2008}]{Lazaric2008}
Alessandro Lazaric, Marcello Restelli, and Andrea Bonarini.
\newblock {Transfer of Samples in Batch Reinforcement Learning}.
\newblock In {\em ICML}, 2008.

\bibitem[\protect\citeauthoryear{Li and Zhang}{2018}]{li2018optimal}
Siyuan Li and Chongjie Zhang.
\newblock {An Optimal Online Method of Selecting Source Policies for
  Reinforcement Learning}.
\newblock In {\em AAAI}, 2018.

\bibitem[\protect\citeauthoryear{Li \bgroup \em et al.\egroup }{2019}]{Li2018}
Siyuan Li, Fangda Gu, Guangxiang Zhu, and Chongjie Zhang.
\newblock {Context-Aware Policy Reuse}.
\newblock In {\em AAMAS}, 2019.

\bibitem[\protect\citeauthoryear{Mankowitz \bgroup \em et al.\egroup
  }{2018}]{mankowitz2018learning}
Daniel~J Mankowitz, Timothy~A Mann, Pierre-Luc Bacon, Doina Precup, and Shie
  Mannor.
\newblock {Learning Robust Options}.
\newblock In {\em AAAI}, 2018.

\bibitem[\protect\citeauthoryear{Osa \bgroup \em et al.\egroup
  }{2019}]{osa2019hierarchical}
Takayuki Osa, Voot Tangkaratt, and Masashi Sugiyama.
\newblock {Hierarchical Reinforcement Learning via Advantage-Weighted
  Information Maximization}.
\newblock In {\em ICLR}, 2019.

\bibitem[\protect\citeauthoryear{Parisotto \bgroup \em et al.\egroup
  }{2016}]{parisotto:actormimic}
Emilio Parisotto, Jimmy Ba, and Ruslan Salakhutdinov.
\newblock Actor-mimic: Deep multitask and transfer reinforcement learning.
\newblock In {\em ICLR}, 2016.

\bibitem[\protect\citeauthoryear{Rajendran \bgroup \em et al.\egroup
  }{2017}]{rajendran:attend}
Janarthanan Rajendran, Aravind~S Lakshminarayanan, Mitesh~M Khapra, P~Prasanna,
  and Balaraman Ravindran.
\newblock Attend, adapt and transfer: Attentive deep architecture for adaptive
  transfer from multiple sources in the same domain.
\newblock In {\em ICLR}, 2017.

\bibitem[\protect\citeauthoryear{Rosman \bgroup \em et al.\egroup
  }{2016}]{rosman2016bayesian}
Benjamin Rosman, Majd Hawasly, and Subramanian Ramamoorthy.
\newblock {Bayesian Policy Reuse}.
\newblock {\em Machine Learning}, 104(1):99--127, 2016.

\bibitem[\protect\citeauthoryear{Schulman \bgroup \em et al.\egroup
  }{2017}]{ppo}
John Schulman, Filip Wolski, Prafulla Dhariwal, Alec Radford, and Oleg Klimov.
\newblock {Proximal Policy Optimization Algorithms}.
\newblock {\em arXiv preprint arXiv:1707.06347}, 2017.

\bibitem[\protect\citeauthoryear{Silver \bgroup \em et al.\egroup
  }{2018}]{Silver2018}
Tom Silver, Kelsey Allen, Josh Tenenbaum, and Leslie Kaelbling.
\newblock {Residual Policy Learning}.
\newblock {\em arXiv preprint arXiv:1812.06298}, 2018.

\bibitem[\protect\citeauthoryear{Song \bgroup \em et al.\egroup
  }{2016}]{Song2016}
Jinhua Song, Yang Gao, Hao Wang, and Bo~An.
\newblock {Measuring the Distance Between Finite Markov Decision Processes}.
\newblock In {\em AAMAS}, 2016.

\bibitem[\protect\citeauthoryear{Sutton and Barto}{1998}]{Sutton1998}
Richard~S. Sutton and Andrew~G. Barto.
\newblock {\em {Introduction to Reinforcement Learning}}.
\newblock MIT Press, 1st edition, 1998.

\bibitem[\protect\citeauthoryear{Sutton \bgroup \em et al.\egroup
  }{1999}]{sutton1999between}
Richard~S Sutton, Doina Precup, and Satinder Singh.
\newblock {Between MDPs and Semi-MDPs: A Framework for Temporal Abstraction in
  Reinforcement Learning}.
\newblock {\em Artificial intelligence}, 112(1-2):181--211, 1999.

\bibitem[\protect\citeauthoryear{Tan \bgroup \em et al.\egroup
  }{2018}]{tan2018}
Jie Tan, Tingnan Zhang, Erwin Coumans, Atil Iscen, Yunfei Bai, Danijar Hafner,
  Steven Bohez, and Vincent Vanhoucke.
\newblock {Sim-to-Real: Learning Agile Locomotion For Quadruped Robots}.
\newblock In {\em RSS}, 2018.

\bibitem[\protect\citeauthoryear{Tange}{2018}]{tange_ole_2018_1146014}
Ole Tange.
\newblock {\em {GNU Parallel 2018}}.
\newblock Ole Tange, 2018.

\bibitem[\protect\citeauthoryear{Tirinzoni \bgroup \em et al.\egroup
  }{2018}]{pmlr-v80-tirinzoni18a}
Andrea Tirinzoni, Andrea Sessa, Matteo Pirotta, and Marcello Restelli.
\newblock {Importance Weighted Transfer of Samples in Reinforcement Learning}.
\newblock In {\em ICML}, 2018.

\bibitem[\protect\citeauthoryear{Vanschoren}{2018}]{metarl}
Joaquin Vanschoren.
\newblock {Meta-Learning: A Survey}.
\newblock {\em arXiv preprint arXiv:1810.03548}, 2018.

\bibitem[\protect\citeauthoryear{Wang \bgroup \em et al.\egroup
  }{2019}]{Wang2019}
Hao Wang, Shaokang Dong, and Ling Shao.
\newblock {Measuring Structural Similarities in Finite MDPs}.
\newblock In {\em IJCAI}, 2019.

\bibitem[\protect\citeauthoryear{Yu \bgroup \em et al.\egroup
  }{2019}]{yu2018policy}
Wenhao Yu, C.~Karen Liu, and Greg Turk.
\newblock {Policy Transfer with Strategy Optimization}.
\newblock In {\em ICLR}, 2019.

\bibitem[\protect\citeauthoryear{Zheng \bgroup \em et al.\egroup
  }{2018}]{zheng2018deep}
Yan Zheng, Zhaopeng Meng, Jianye Hao, Zongzhang Zhang, Tianpei Yang, and
  Changjie Fan.
\newblock {A Deep Bayesian Policy Reuse Approach against Non-Stationary
  Agents}.
\newblock In {\em NeurIPS}, 2018.

\end{thebibliography}

\end{document}